\documentclass{article}

    \usepackage[nonatbib,final]{neurips_2022}

\usepackage[utf8]{inputenc} % allow utf-8 input
\usepackage[T1]{fontenc}    % use 8-bit T1 fonts
\usepackage{hyperref}       % hyperlinks
\usepackage{url}            % simple URL typesetting
\usepackage{booktabs}       % professional-quality tables
\usepackage{amsfonts}       % blackboard math symbols
\usepackage{nicefrac}       % compact symbols for 1/2, etc.
\usepackage{microtype}      % microtypography
\usepackage{xcolor}         % colors
\usepackage{multirow}
\usepackage{subfigure}
\usepackage{enumitem}
\usepackage{graphicx}
\usepackage{wrapfig,lipsum}
\usepackage{amssymb,amsmath}

\usepackage{arydshln}

\title{Augmentations in Hypergraph Contrastive Learning: Fabricated and Generative}
\author{%
  Tianxin Wei$^{1*}$, Yuning You$^{2*}$, Tianlong Chen$^3$, Yang Shen$^2$, Jingrui He$^1$, Zhangyang Wang$^3$\\
  $^1$University of Illinois Urbana-Champaign, $^2$Texas A\&M University,
  $^3$University of Texas at Austin\\
  \texttt{\{twei10,jingrui\}@illinois.edu}, \texttt{\{yuning.you,yshen\}@tamu.edu},\\ \texttt{\{tianlong.chen,atlaswang\}@utexas.edu}\\
}

\begin{document}

\maketitle

\begin{abstract}
This paper targets at improving the generalizability of hypergraph neural networks in the low-label regime, through applying the contrastive learning approach from images/graphs (we refer to it as \textbf{HyperGCL}).
We focus on the following question: \textit{How to construct contrastive views for hypergraphs via augmentations?}
We provide the solutions in two folds.
First, guided by domain knowledge, we \textbf{fabricate} two schemes to augment hyperedges with higher-order relations encoded, and adopt three vertex augmentation strategies from graph-structured data.
Second, in search of more effective views in a data-driven manner, we for the first time propose a hypergraph generative model to \textbf{generate} augmented views, and then an end-to-end differentiable pipeline to jointly learn hypergraph augmentations and model parameters.
Our technical innovations are reflected in designing both fabricated and generative augmentations of hypergraphs.
The experimental findings
include:
(i) Among fabricated augmentations in HyperGCL, augmenting hyperedges provides the most numerical gains, implying that higher-order information in structures is usually more downstream-relevant;
(ii) Generative augmentations do better in preserving higher-order information to further benefit generalizability;
(iii) HyperGCL also boosts robustness and fairness in hypergraph representation learning.
Codes are released at \url{https://github.com/weitianxin/HyperGCL}.

% \he{What are numerical findings?}

% --------------------------------------------------------

% Hypergraph, a flexible structure for modeling higher-order correlations among objects, has recently gained increased interest from a variety of academic disciplines, and there are numerous practical hypergraph datasets available. However, generalizable, robust, and fair representation learning on hypergraphs remains a challenge for current hypergraph neural networks. In this paper, we introduce contrastive self-supervision into the hypergraph, to provide general-purpose self-supervised representation for the first time. Moreover, we present the first thorough investigation and evaluation to show when and how contrastive self-supervision help HyperGCNs from five perspectives: mechanism, augmentation, model, data, and task. We conduct extensive experiments on both the existing benchmark and the newly curated fair and robust datasets to demonstrate that, with properly designed augmentations and incorporation mechanisms, contrastive self-supervision benefits HyperGCNs in gaining more generalizability, robustness, and fairness compared to state-of-the-art methods. We also make the first attempt to explore automated augmentation in the hypergraph domain and have some valuable lessons.
\end{abstract}

\renewcommand{\thefootnote}{\fnsymbol{footnote}}
\footnotetext[1]{Equal contribution.}
\renewcommand{\thefootnote}{\arabic{footnote}}

\section{Introduction} \label{sec:introduction}

Hypergraphs  
%which are generalized from ordinary graph-structured data \cite{kipf2016semi},   
have raised a surge of interests in the research community \cite{feng2019hypergraph,yadati2019hypergcn,chien2021you} due to their innate capability of capturing higher-order relations \cite{bretto2013hypergraph}.
They offer a powerful tool to model complicated topological structures in broad applications, e.g., recommender systems \cite{ji2020dual, yu2021self}, financial analyses \cite{sawhney2020spatiotemporal,you2022cross}, and bioinformatics \cite{zhang2022multiscale,you2022cross,liu2022generating}.
Concomitant with the trend, hypergraph neural networks (HyperGNNs) have recently been developed \cite{feng2019hypergraph,yadati2019hypergcn,chien2021you} for hypergraph representation learning.

This paper focuses on the few-shot scenarios of hypergraphs, i.e., task-specific labels are scarce, which are ubiquitous in real-world applications of hypergraphs \cite{ji2020dual,sawhney2020spatiotemporal,zhang2022multiscale} and empirically restrict the generalizability of HyperGNNs.
Inspired by the emerging self-supervised learning on images/graphs \cite{goyal2019scaling,chen2020simple,hu2019strategies,you2020graph,you2020does,jin2020self}, especially the contrastive approaches \cite{chen2020simple,you2020graph,xie2022self,velickovic2019deep,you2021graph,you2022bringing,zhu2020deep,trivedi2022augmentations,wang2022augmentation,feng2022adversarial,xu2022group}, we set out to leverage contrastive self-supervision to address the problem.

Nevertheless, one challenge stands out: \textit{How to build contrastive views for hypergraphs?}
The success of contrastive learning hinges on the appropriate view construction, otherwise it would result in ``negative transfer'' \cite{chen2020simple,you2020graph}.
However, it is non-trivial to build hypergraph views due to their overly intricate topology, i.e., there are $\sum_{e=1}^N \bigl( \begin{smallmatrix} N \\ e \end{smallmatrix} \bigr)$ possibilities for one hyperedge on $N$ vertices, versus $\bigl( \begin{smallmatrix} N \\ 2 \end{smallmatrix} \bigr)$ for one edge in graphs.
To date, the only way of contrasting is between the representations of hypergraphs and their clique-expansion graphs
% \tx{tx: their original data is graph, the hypergraph is generated by transformation}
\cite{xia2022hypergraph,cai2022hypergraph}, which is computationally expensive as multiple neural networks of different modalities (hypergraphs and variants of expanded graphs) need to be optimized.
More importantly, contrasting on clique expansion has the risk of losing higher-order information via pulling representations of hypergraphs and graphs close.

\textbf{Contributions.}
Motivated by \cite{chen2020simple,you2020graph} that appropriate data augmentations suffice for the effective contrastive views, and intuitively they are more capable of preserving  higher-order relations in hypergraphs compared to clique expansion, we explore on the question in this paper, how to design augmented views of hypergraphs in contrastive learning (\textbf{HyperGCL}).
Our answers are in two folds.

We first assay whether \textbf{fabricated} augmentations guided by domain knowledge are suited for HyperGCL.
Since hypergraphs are composed of hyperedges and vertices,
to augment hyperedges, we propose two strategies that (i) directly perturb on hyperedges, and (ii) perturb on the ``edges'' between hyperedges and vertices in the converted bipartite graph;
To augment vertices, we adopt three schemes of vertex dropping, attribute masking and subgraph from graph-structured data \cite{you2020graph}. Our finding is that, different from the fact that vertex augmentations benefit more on graphs, \textit{hypergraphs mostly benefit from hyperedge augmentations} (up to 9\% improvement), revealing that higher-order information encoded in hyperedges is  usually more downstream-relevant (than information in vertices).

% We adopt vertices augmentations of vertex dropping, attribute masking and subgraph from GraphCL \cite{you2020graph}.
% To augment hyperedges, we propose two strategies, that (i) augmenting on the "edges" between vertices and hyperedges of the converted bipartite graph, and (ii) directly augmenting on hyperedges.
% Our finding is, different from the fact that vertex augmentations are generally beneficial for graphs, \textit{hypergraphs are mostly boosted with hyperedge augmentations}.
% This coincides to the nature that, higher-order information encoded in hyperedges are usually more downstream-relevant than that in vertices.

Furthermore, in search of even better augmented views but in a data-driven manner, we study whether/how augmentations of hypergraphs could be learned during contrastive learning.
To this end, for the first time, we propose a novel variational hypergraph auto-encoder architecture, as a hypergraph \textbf{generative} model, to parameterize a certain augmentation space of hypergraphs.
In addition, we propose an end-to-end differentiable pipeline utilizing Gumbel-Softmax \cite{jang2016categorical}, to jointly learn hypergraph augmentations and model parameters.
Our observation is that generative augmentations can better capture the higher-order information and achieve state-of-the-art performance on most of the benchmark data sets (up to 20\% improvement).

The aforementioned empirical evidences (for \textit{generalizability}) are drawn from comprehensive experiments on 13 datasets.
Moreover, we introduce the \textit{robustness} and \textit{fairness} evaluation for hypergraphs, and show that HyperGCL in addition boosts robustness against adversarial attacks and imposes fairness with regard to sensitive attributes.

The rest of the paper is organized as follows. We discuss the related work in Section \ref{section:related work}, introduce HyperGCL in Section \ref{section:method}, present the experimental results in Section \ref{section:experiments}, and conclude in Section \ref{section:conclusion}.

\section{Related Work}
\label{section:related work}
% \yy{(Todo: rewrite this section)}
\textbf{Hypergraph neural networks.} 
Hypergraphs, which are able to encode higher-order relationships, have attracted significant attentions in recent years. In the machine learning community, hypergraph neural networks are developed for effective hypergraph representations.
% A natural solution for learning on hypergraphs is to transform a hypergraph into a conventional graph for easy adaptation. Clique expansion technique is widely used to convert the hypergraph.
HGNN \cite{feng2019hypergraph} adopt the clique expansion technique and designs the weighted hypergraph Laplacian for message passing. HyperGCN \cite{yadati2019hypergcn} proposes the generalized hypergraph Laplacian and explores adding the hyperedge information through mediators. The attention mechanism \cite{zhang2019hyper,bai2021hypergraph} is also designed to learn the importance within hypergraphs. However, the expanded graph will inevitably cause distortion and lead to unsatisfactory performance. There is also another line of works such as UniGNN \cite{huang2021unignn} and HyperSAGE \cite{arya2020hypersage} which try to perform message passing directly on the hypergraph to avoid the information loss. A recent work \cite{chien2021you} provides an AllSet framework to unify the existing studies with high expressive power and achieves state-of-the-art performance on comprehensive benchmarks. The work utilizes deep multiset functions \cite{zaheer2017deep} to identify the propagation and aggregation rules in a data-driven manner.
% However, few work focuses on the few-shot scenario, which is limited in task-specific labels.
% In this paper, we generate more effective hypergraph representation by first exploring novel contrastive self-supervised strategies.

%LEGCN first transforms hypergraphs into graphs via line expansion and then applies a standard GCN. Although the transformation used in LEGCN does not
% cause a large distortion as CE, it significantly increases the number of vertexes and edges. Hence, it is not memory and computationally efficient.

\textbf{Contrastive self-supervised learning.} 
% \cite{du2021hypergraph} designs pretext tasks for hypergraphs based on clustering. But, it doesn't enjor the modeling power of contrastive leanring and focuses on hyperedge prediction while we focus on semi-supervised vertex classification. 
Contrastive self-supervision \cite{chen2020simple,he2020momentum,grill2020bootstrap} has achieved unprecedented success in computer vision. The core idea is to learn an embedding space where samples from the same instance are pulled closer and samples from different instances are pushed apart. Recent works start to cross-pollinate between contrastive learning and graph neural networks to for more generalizable graph representations. Typically, they design some fabricated augmentations guided by domain knowledge, such as edge perturbation, feature masking or vertex dropping, etc. 
%(YS: ignored the body of work on generative views for graphs)
% With the rapid development of graph contrastive learning, self-supervised
Nevertheless, contrastive learning on hypergraphs remains largely unexplored. Most existing works \cite{yu2021self,xia2021self,xia2022hypergraph,du2021hypergraph} design pretext tasks for hypergraphs and mainly focus on recommender systems \cite{he2017neural, wei2021model,wei2022comprehensive,zhang2021causal}, via contrasting between graphs and hypergraphs which might lose important higher-order information.
% Moreover, they solve a very different problem by generating hypergraph from the conventional graph based on the characteristics of the data. The hypergraph only serves as a complementary function.
In this work, we explore on the structure of hypergraph itself to construct contrastive views.

\section{Methods}
\label{section:method}

\begin{figure}[t] 
\centering
\includegraphics[width=0.8\linewidth]{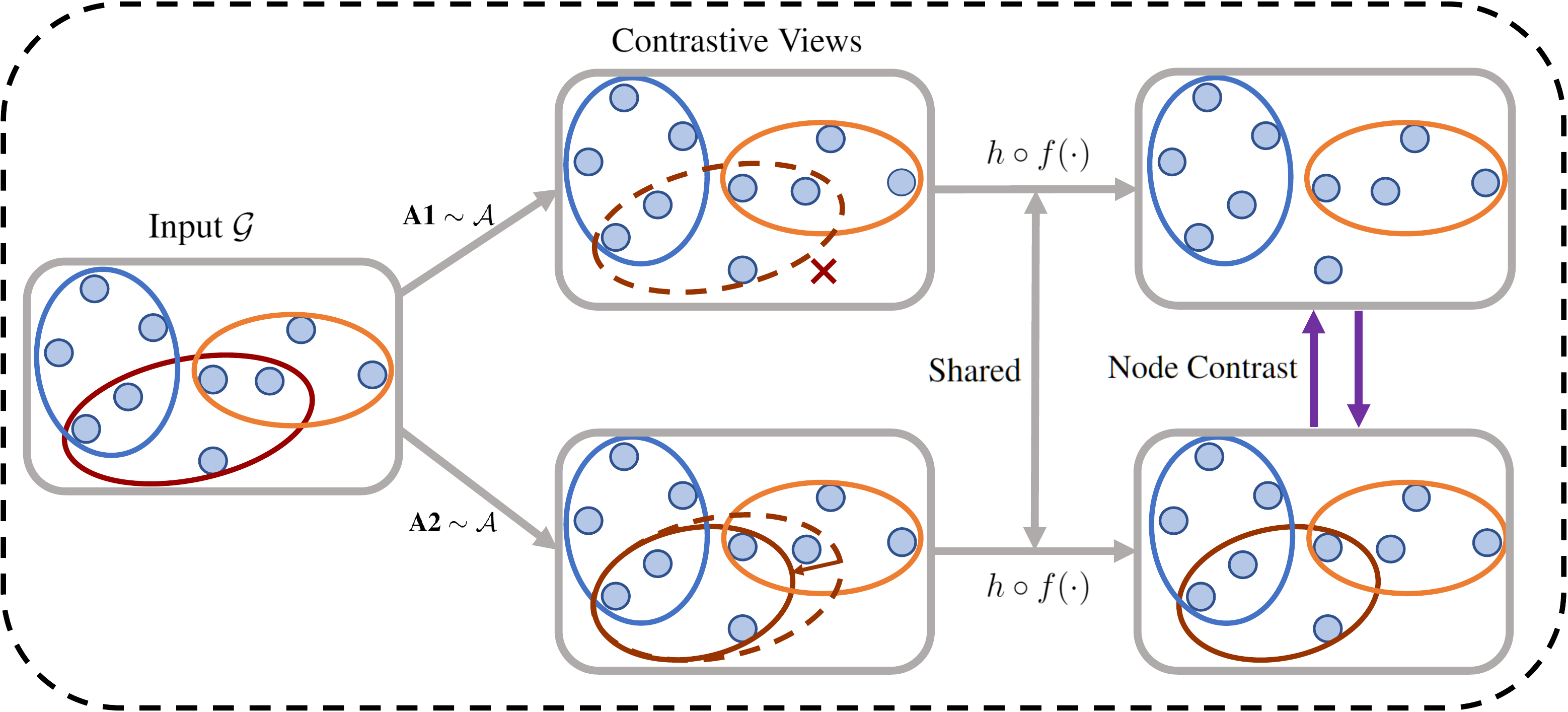}
\caption{The framework of hypergraph contrastive learning (HyperGCL). The ellipses represent the hyperedges. Two contrastive views are generated by hypergraph augmentations \textbf{A1} and \textbf{A2} from the augmentation collection $\mathcal{A}$. $f(\cdot)$ and $h(\cdot)$ are shared encoder and projection head respectively. In the figure, we show two examples of hypergraph augmentations. At the top, the dotted ellipse denotes the deleted hyperedge. At the bottom, one vertex in the dotted hyperedge is removed.}
% \vspace{-0.3cm}
\label{fig:frame}
% \vspace{-0.6cm}
\end{figure}
\subsection{Hypergraph Contrastive Learning}
A hypergraph is denoted as $\mathcal{G} = \{ \mathcal{V},\mathcal{E} \} \in \mathbb{G}$ where $\mathcal{V}=\{v_1,...,v_{|\mathcal{V}|}\}$ is the set of vertices and $\mathcal{E}=\{e_1,...,e_{|\mathcal{E}|}\}$ is the set of hyperedges. Each hyperedge $e_n = \{ v_1,...,v_{|e_n|} \}$ represents the higher-order interaction among a set of vertices.
State-of-the-art approaches to encode such complex structures are hypergraph neural networks (HyperGNNs) \cite{feng2019hypergraph,yadati2019hypergcn,chien2021you}, mapping the hypergraph to a $D$-dimension latent space via $f: \mathbb{G} \rightarrow \mathbb{R}^D$ with higher-order message passing.

Motivated from learning on images/graphs, we adopt contrastive learning to further improve the generalizability of HyperGNNs in the low-label regime (HyperGCL).
Main components of our HyperGCL, similar to images/graphs \cite{chen2020simple,you2020graph} include:
(i) \textbf{hypergraph augmentations for contrastive views}, (ii) HyperGNNs as hypergraph encoders, (iii) projection head $h(\cdot)$ for representations, and (iv) contrastive loss for optimization.
The overall pipeline is shown in Figure \ref{fig:frame}.
% $\mathcal{L}_{cl}$ denotes the contrastive loss.
Detailed descriptions and training procedure are shown in Appendix B.
The main challenge here is how to effectively augment hypergraphs to build contrastive views.

% Given a hypergraph $\mathcal{G}$, two contrastive views $\hat{\mathcal{G}}_1$ and $\hat{\mathcal{G}}_2$ are first generated by hypergraph data augmentations $\mathcal{T}$. The vertex embedding for each hypergraph view can be computed as $Z^\mathcal{V}_{1}=f(\hat{\mathcal{G}}_1)$, $Z^\mathcal{V}_{2}=f(\hat{\mathcal{G}}_2)$. In the two hypergraph views, the corresponding vertex pairs are positive pairs, while all other vertex pairs are denoted as negative. Given the cosine similarity function as $\gamma(\mathbf{u},\mathbf{s})=\frac{\mathbf{u}^T\mathbf{s}}{||\mathbf{u}|| ||\mathbf{s}||}$, $\mathbf{u}_n=h(Z^\mathcal{V}_{1})[n,:]$ and $\mathbf{s}_n=h(Z^\mathcal{V}_{2})[n,:]$, the contrastive loss is constructed as:
% \vspace{-0.2cm}
% \begin{align}
% \label{eq:contra}
% \mathcal{L}_{cl}(\hat{\mathcal{G}}_i,\hat{\mathcal{G}}_j) = \frac{1}{2|\mathcal{V}|}\sum_{i=1}^{|\mathcal{V}|}(l(\mathbf{u}_n,\mathbf{s}_n)+l(\mathbf{s}_n,\mathbf{u}_n)),
% \end{align}
% \vspace{-0.2cm}
% \begin{align}
% l(\mathbf{u}_n,\mathbf{s}_n) =
% -\log{\frac{e^{\gamma(\mathbf{u}_n,\mathbf{s}_n)/\tau}}{e^{\gamma(\mathbf{u}_n,\mathbf{s}_n)/\tau} +\sum\limits_{m\neq n}e^{\gamma(\mathbf{u}_n,\mathbf{s}_m)/\tau}+\sum\limits_{m\neq n}e^{\gamma(\mathbf{u}_n,\mathbf{u}_m)/\tau}}},
% \end{align}
% where $\tau$ is a temperature parameter. The overall pipeline is depicted in Figure \yy{(\ref{fig:frame})}.

% The loss $l$ is symmetrically defined. In principle, our framework could be applied on any HyperGCN architectures.

\subsection{Fabricated Augmentations for Hypergraphs} \label{sec:fabricated_augmentations}

\begin{wrapfigure}{r}{0.7\textwidth}
\begin{center}
\includegraphics[width=0.7\linewidth]{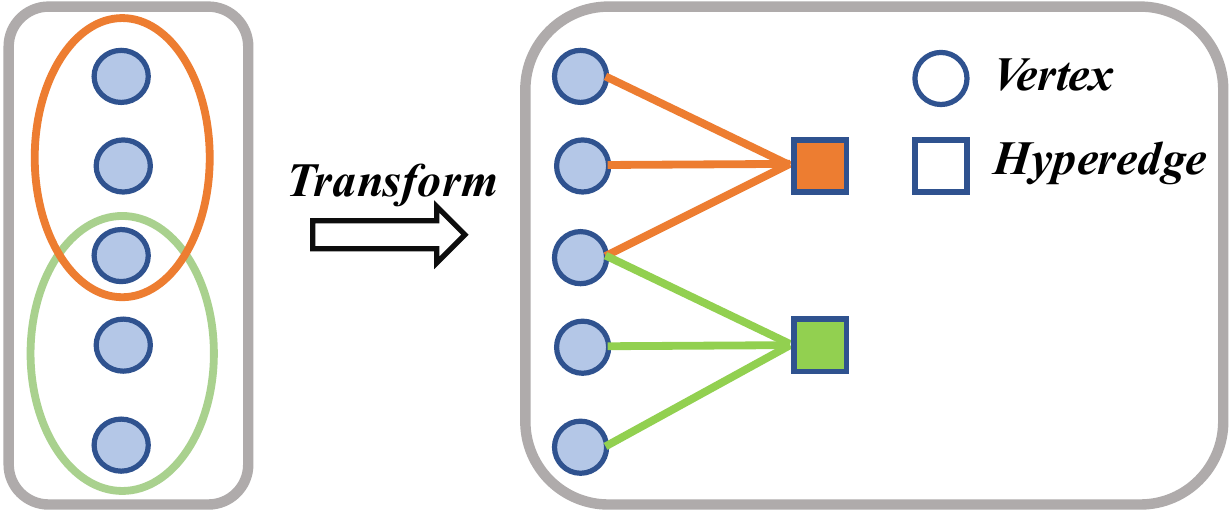}
\end{center}
\caption{Conversion from hypergraph to equivalent bipartite graph.}
% \vspace{-0.3cm}
\label{fig:transform}
% \vspace{-0.4cm}
\end{wrapfigure}
We first explore whether manually designed augmentations are suited for HyperGCL.
Since hyperedges and vertices compose a hypergraph, augmentations are fabricated with regards to topology and node features, respectively.

% Augmentations are designed with regard to hyperedges and vertices, respectively, which compose the hypergraph.\he{I don't see the discussion of domain knowledge in this subsection.}

\textbf{A1. Perturbing hyperedges.}
The most direct augmentation on higher-order interactions is to perturb on the set of hyperedges.
Since adding a hyperedge is confronted with the combinatorial challenge (see Sec. \ref{sec:introduction} of introduction), here we focus on randomly removing the existing hyperedges following an i.i.d. Bernoulli distribution.
The underlying assumption is that the partially missing higher-order relations do not significantly affect the semantic meaning of hypergraphs.

\textbf{A2. Perturbing edges in equivalent bipartite graph.}
To augment higher-order relations in a more fine-grained way, we first convert the hypergraph into the equivalent bipartite graph, where two disjoint sets of vertices represent vertices and hyperedges in the hypergraph, respectively (see Figure \ref{fig:transform}).
On top of the bipartite graph, we perform random removal of edges.
A2 disrupts the higher-order relations via randomly kicking out vertices from hyperedges, enforcing the semantics of hypergraph representations to be robust to such disruption.
A2 is essentially the generalized version of A1.
% \vspace{-0.3cm}

Moreover, we find that vertex augmentations for  graph-structured data \cite{you2020graph} are applicable to hypergraphs. Therefore, we adopt three additional schemes of vertex dropping (\textbf{A3}), attribute masking (\textbf{A4}) and subgraph (\textbf{A5}) into our experiments, with similar prior knowledge incorporated as in \cite{you2020graph}.

\subsection{Generative Models for Hypergraph Augmentations}
Manually designing augmentation operators requires a wealth of domain knowledge, and might lead to sub-optimal solutions even with extensive trial-and-errors.
We next study whether/how augmentations of hypergraphs could be learned during contrastive learning.
Two questions need to be answered here: (i) How to parameterize the augmentation space of hypergraphs? (ii) How to incorporate the learnable augmentations into contrastive learning?

\subsubsection{Hypergraph Generative Models for Augmentations}
Considering an augmentation operator defined as the stochastic mapping between two hypergraph manifolds that $g: \mathbb{G} \rightarrow \mathbb{G}$, a natural thought is to adopt the generative model to parameterize the augmentation space, which in general is composed of a deterministic encoder $h_1: \mathbb{G} \rightarrow \mathbb{R}^{D'}$ and a stochastic decoder (or sampler) $h_2: \mathbb{R}^{D'} \rightarrow \mathbb{G}$. In this way, $g = h_1 \circ h_2$.

\begin{figure}[t] 
\centering
\includegraphics[width=0.9\linewidth]{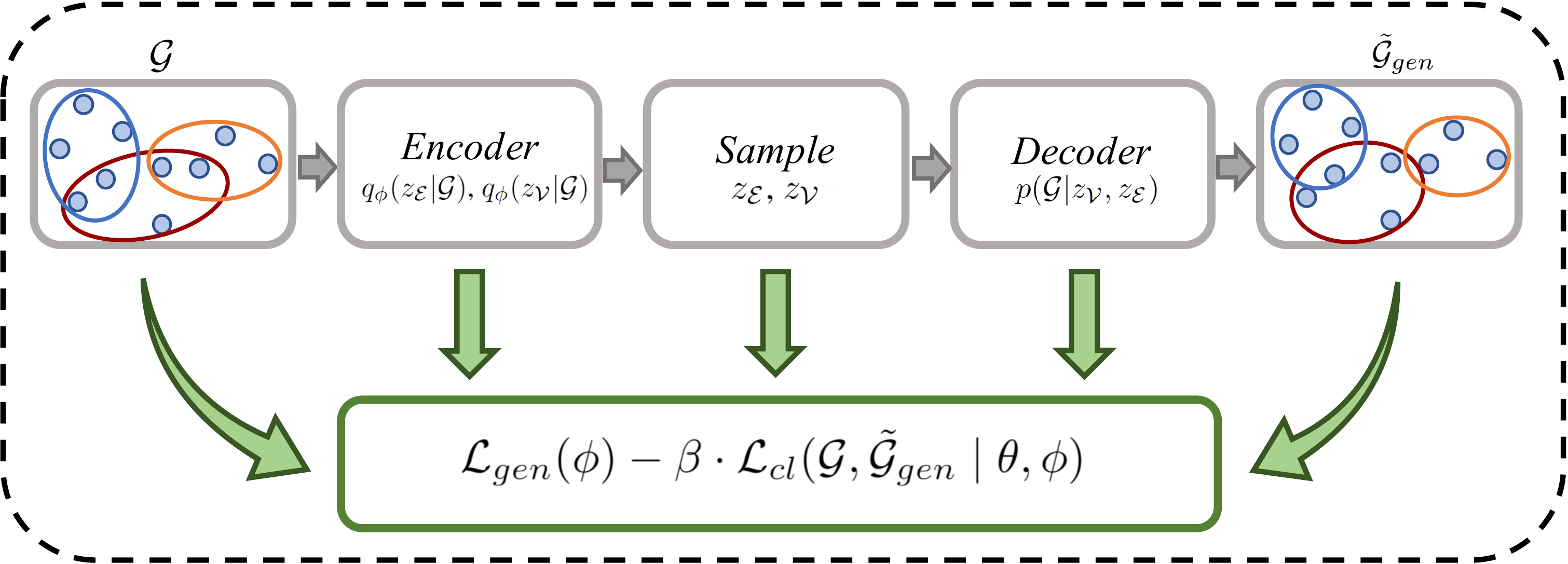}
\caption{Framework of the proposed variational hypergraph auto-encoder (VHGAE).
The green lines indicate these modules participated in the optimization process.
}
% \vspace{-0.3cm}
\label{fig:vhgae}
% \vspace{-0.6cm}
\end{figure}

Following this thought, inspired by the well-studied generative models with variational inference \cite{kingma2013auto,kipf2016variational}, we propose a novel variational hypergraph auto-encoder architecture (VHGAE). To the best of our knowledge, this is the first hypergraph generative model for generating augmentations of hypergraphs, which will be used as \textbf{A6}.
Notice that here it only parametrizes the augmentation space of edge perturbation, and in the future node perturbation would be included.
% We choose VAE-like generative model because the hypergraph generation process is not fixed. Different graph structure can correspond to the same prediction results. With VAE-like model, we can enjoy the high expressive power as well as the stochastic variance.
% Following convention,
VHGAE consists of the encoder and decoder neural networks. The overall framework is shown in Figure \ref{fig:vhgae}.

\textbf{Encoder.}
The encoder embeds hypergraphs into latent representations.
Instead of embedding a hypergraph into a single vector, we follow VGAE \cite{kipf2016variational} to embed it into a set of vertex and in additional hyperedge representations, to facilitate the further decoding process of non-Euclidean structures.
We adopt two HyperGNNs, $h_1^\mu \text{ and } h_1^\sigma$, to encode the mean and the logarithmic standard deviation for variational distributions of vertex and hyperedge representations $z_\mathcal{V} \sim q_{\phi}(z_{\mathcal{V}}|\mathcal{G})=\mathcal{N}(\mu_{\mathcal{V}}, \sigma_{\mathcal{V}}^2), z_\mathcal{E} \sim q_{\phi}(z_{\mathcal{E}}|\mathcal{G})=\mathcal{N}(\mu_{\mathcal{E}}, \sigma_{\mathcal{E}}^2)$ as follows (please refer to Appendix B for the detailed computing pipeline):
\begin{equation}
    \mu_{\mathcal{V}}, \mu_{\mathcal{E}} = h_1^\mu(\mathcal{G}), \quad\quad \log(\sigma_{\mathcal{V}}), \log(\sigma_{\mathcal{E}}) = h_1^\sigma(\mathcal{G}),
\end{equation}
where $\mu \in \mathbb{R}^{D' \times |\mathcal{V}|}, \log(\sigma) \in \mathbb{R}^{D' \times |\mathcal{V}|}$.
% , and thereby $z_\mathcal{V} \sim \mathcal{N}(\mu_{\mathcal{V}}, \sigma_{\mathcal{V}}^2), z_\mathcal{E} \sim \mathcal{N}(\mu_{\mathcal{E}}, \sigma_{\mathcal{E}}^2)$
% \he{What are these?}.
We here leverage the higher-order message passing in HyperGNNs for a better encoding capability.

\textbf{Decoder.}
With the learned vertex and hyperedge variational distributions, the decoder attempts to reconstruct the higher-order relations of hypergraphs.
However, modeling the space of higher-order interactions
% (for\he{What's this?})
encounters the combinatorial challenge (see Section \ref{sec:introduction}).
Adopting the similar strategy as in the augmentation A2 (see Section \ref{sec:fabricated_augmentations}), we designate the decoder to recover the relations on the converted bipartite graph $\tilde{\mathcal{G}} = \{\tilde{\mathcal{V}}, \tilde{\mathcal{E}}\}$ for approximation.
Mathematically, we formulate decoding as:
\begin{equation}
    p(\mathcal{G}|z_{\mathcal{V}},z_{\mathcal{E}}) \approx p(\tilde{\mathcal{G}}|z_{\mathcal{V}},z_{\mathcal{E}}) = \prod_{e=1}^{|\mathcal{E}|} \prod_{v=1}^{|\mathcal{V}|} p(\tilde{\mathcal{E}}_{v,e}|z_v,z_e) = \prod_{e=1}^{|\mathcal{E}|} \prod_{v=1}^{|\mathcal{V}|} \mathrm{Sigmoid} (z_v^T z_e),
\end{equation}
where $w_{ve}=z_v^T z_e$ is the learned edge logit. On the decoded topological distribution of the bipartite graph, we perform sampling and then convert the sample back to the hypergraph
(the conversion between hypergraphs and bipartite graphs is lossless).

\textbf{Generator optimization.}
With variational inference \cite{kingma2013auto,kingma2019introduction,you2021bayesian}, we optimize the hypergraph generator on the evidence lower bound (ELBO) as follows:
% \begin{small}
\begin{align}
\text{ELBO}=\mathbb{E}_{q_{\phi}(z_{\mathcal{E}} \mid \mathcal{G})} \mathbb{E}_{q_{\phi}(z_{\mathcal{V}} \mid \mathcal{G})}\left[\log p_{\theta}(\mathcal{G}|z_v,z_e)\right]-\text{KL}[q_{\phi}(z_{\mathcal{V}} \mid \mathcal{G})\mid p(z_{\mathcal{V}})]-\text{KL}[q_{\phi}(z_{\mathcal{E}} \mid \mathcal{G})\mid p(z_{\mathcal{E}})],
\end{align}
% \end{small}
where $q_{\phi}(z_{\mathcal{E}}|\mathcal{G})$ and $q_{\phi}(z_{\mathcal{V}}|\mathcal{G})$ are their variational distribution, $p(z_{\mathcal{V}})$ and $p(z_{\mathcal{E}})$ are default Gaussian priors with $p(z_{\mathcal{V}})\sim \mathcal{N}(0,I)$, $p(z_{\mathcal{E}})\sim \mathcal{N}(0,I)$. When generating hypergraphs, the generator would  sample the relations on the converted bipartite graph with probability $p(\tilde{\mathcal{G}}|z_{\mathcal{V}},z_{\mathcal{E}})$.

% \begin{align}
% \begin{aligned}
%     &p(\tilde{\mathcal{G}}|z_{\mathcal{V}},z_{\mathcal{E}})=\prod_{e=1}^{|\mathcal{E}|} \prod_{v=1}^{|\mathcal{V}|} p(\tilde{\mathcal{E}}_{v,e}|z_v,z_e),~\text{with}~ p(\tilde{\mathcal{E}}_{v,e}|z_v,z_e)=\sigma(z_v^Tz_e),
% \end{aligned}
% \end{align}

\subsubsection{Jointly Augmenting and Contrasting with Gumbel-Softmax}
With  hypergraph augmentations  parametrized with generative models, the next step is to incorporate augmentation learning into HyperGCL.
The main barrier results from the discrete sampling of hyperedges which is non-differentiable. To tackle it, we leverage the Gumbel-Softmax trick \cite{jang2016categorical} for the hyperedge distribution as:
\begin{align}
\begin{aligned}
        T(\mathcal{G})&=\text{Gumbel-Softmax}(p(\mathcal{G}\mid z_{\mathcal{V}},z_{\mathcal{E}})) \\
        &=\text{Sigmoid}((w_{\mathcal{V}\mathcal{E}}+\text{log}(\delta)-\text{log}(1-\delta))/\tau)\\
        \tilde{\mathcal{G}}_{gen}&=T(\mathcal{G})\circ \mathcal{G},
\end{aligned}
\end{align}
where $w_{\mathcal{V}\mathcal{E}}$ denotes the learned edge logits (before Sigmoid) and $\delta\sim \text{Uniform}(0,1)$. When hyperparameter temperature $\tau\rightarrow 0$, the results get closer to being binary. $T$ is the sampled one-hot vector for each hyperedge-vertex interaction in the hypergraph $\mathcal{G}$. Then the sampled vector will be applied to perform augmentation. During the Gumbel-Softmax, we leverage the reparametrization trick \cite{kingma2013auto} to smooth the gradient and make the sample operation differentiable. Thus, this objective can be optimized in an end-to-end manner as:
\begin{align}
\begin{aligned}
    \text{min}_{\phi} \mathcal{L}_{gen}(\phi)-\beta\cdot \mathcal{L}_{\text{cl}}(\mathcal{G},\tilde{\mathcal{G}}_{gen}\mid \theta,\phi),
\end{aligned}
\end{align}
where $\mathcal{L}_{gen}=-\text{ELBO}$ is the generator loss to be minimized, $\beta$ is the tradeoff factor.
Due to the computational cost of collaboratively optimizing two generative views, we train one VHGAE to produce one generative view, with the other view $\tilde{\mathcal{G}}_p$ is kept as fabricated. To be specific, (i) independently optimizing two hypergraph generators is of reasonable budgets but would lead to distribution collapse (i.e., two hypergraph generators output the same distribution) [1,2] which results in less effective generative views, while (ii) the collaborative optimization techniques for graph generators (e.g. REINFORCE on the rewards of generative graph structures) are not directly applicable to HyperGCL due to the combinatorial challenge of hypergraph structures (which is computationally expensive). The goal of this multi-task loss is to generate stronger augmentation (maximize contrastive loss) to push HyperGNN to avoid capturing redundant information during the representation learning, while at the same time learning the hypergraph data distribution.

\section{Experiments}
\label{section:experiments}

\begin{table}[ht]
\centering
\caption{Data statistics: $h_e$, $h_v$ are the node homophily and hyperedge homophily in hypergraph. Higher value indicate the hypergraph is more homogeneous. Details can be found in Appendix C.}
\label{tab:datastats}
\setlength{\tabcolsep}{2.5pt}
\scriptsize
\begin{tabular}{@{}cccccccccccccc@{}}
\toprule
            & Cora & Citeseer & Pubmed & Cora-CA & DBLP-CA & Zoo   & 20News & Mushroom & NTU2012 & ModelNet40 & Yelp  & House & Walmart \\ \midrule
$|\mathcal{V}|$    & 2708    & 3312        & 19717     & 2708    & 41302   & 101   & 16242  & 8124     & 2012    & 12311      & 50758 & 1290  & 88860   \\
$|\mathcal{E}|$           & 1579 & 1079 & 7963 & 1072 & 22363 & 43 & 100  & 298  & 2012 & 12311 & 679302 & 341  & 69906 \\
$\#$  feature        & 1433 & 3703 & 500  & 1433 & 1425  & 16 & 100  & 22   & 100  & 100   & 1862   & 100    & 100    \\
$\#$  class          & 7    & 6    & 3    & 7    & 6     & 7  & 4    & 2    & 67   & 40    & 9      & 2    & 11    \\
$h_e$     & 0.86   &  0.83 &  0.88      &    0.88     &     0.93   & 0.66 &    0.73 &  0.96 &  0.87    &  0.92    & 0.57 & 0.58       &       0.75 \\
$h_v$  & 0.84    & 0.78 &  0.79   &       0.79 &          0.88 &   0.35    &  0.49  & 0.87 &    0.81  &     0.88  &  0.26 &              0.52     &         0.55
\\ \bottomrule
\end{tabular}
% \vspace{-0.2cm}
\end{table}

\subsection{Setup}
We examine our methods on the most comprehensive hypergraph benchmarks with 13 data sets, with statistics shown in Table \ref{tab:datastats}.
Please refer to Appendix C for detailed information.
We focus on semi-supervised vertex classification in the transductive setting. Different from the existing work \cite{chien2021you} that leverages 50\% of all vertexes as the training set, we focus on the more low-label regime of challenging and practical applications.
% , which is more consistent with the real-world applications.
By default, we split the data into training/validation/test samples using (10\%/10\%/80\%) splitting percentages.
Each experiment is run for 20 different data splits and initialization with mean and standard deviation reported.
% We collect the results of 20 experiments using multiple random splits and initializations.
We adopt state-of-the-art SetGNN \cite{chien2021you}
as the backbone HyperGNN architecture.
% with the default setting as the HyperGNN-based encoder which is SOTA in representation learning.
For baselines, we compare two existing hypergraph self-supervised approaches \cite{xia2021self} and \cite{xia2022hypergraph} in recommender systems, denoted as Self and Con. They conduct self-supervised learning between the hypergraph and conventional graph. By default, we adopt multi-task training to incorporate contrastive self-supervision because it performs the best as shown in the comparison in Section \ref{subsec:results}. All the implementation details are listed in Appendix C. More experiments of the hyperparameters study are given in Appendix A.
% In their approaches, the authors construct the hypergraph based on recommended data, and compare the learned embeddings between the hypergraphs and graphs to conduct self-supervised or contrastive learning. We adapt their idea into general hypergraph learning, and perform self-supervised learning between the hypergraph and the clique-expanded graph. Our generative method is named as HyperGCL.

\subsection{Results}
\label{subsec:results}
\begin{wraptable}{r}{6.7cm}
\caption{The proposed augmentation operations for contrastive learning framework HyperGCL and their corresponding names.}
\label{tab:operations}
\scalebox{0.85}{
\begin{tabular}{cl}
\toprule
Name & Operation              \\ \midrule
A0   & Identity               \\
A1   & Na\"ive Hyperedge Perturbation \\
A2   & Generalized Hyperedge Perturbation      \\
A3   & Vertex Dropping        \\
A4   & Attribute Masking      \\
A5   & Subgraph               \\ \hdashline
A6   & Generative Augmentation               \\ \bottomrule
\end{tabular}}
\end{wraptable}
\textbf{Comparison among different hypergraph augmentations.}
% \he{Consider using A1, ..., A6 to connect with the earlier part of your paper.}
The augmentation operations are summarized in Table \ref{tab:operations}. Please refer to Appendix C.5 for detailed descriptions. We first conduct experiments to compare different contrastive operations on hypergraphs, with results shown in Table \ref{tab:10Results}.
In general, generalized hyperedge augmentation (A2) works the best among fabricated augmenting operators, but not na\"ively perturbing hyperedge (A1).
Specifically, among all fabricated augmentations, A2 performs the best in 10 of 13 data sets.
This indicates the nature that higher-order information in structures is usually more downstream-relevant.

% For the base model SetGNN, we find the performance is strongly related to the hypergraph homophily.
% Then for fabricated augmentations, we find Edge performs the best in general. Edge method can perturb the connectivity of hyperedges in a fine granularity, while hyperedge and vertex drop will directly delete the whole hyperedge or vertex, which may have substantial influence on the model. When considering the Mask approach, it can only perturb the vertex features while losing the rich information in the hypergraph.

For our generative augmentation (A6), we find it performs the best in all the data sets. In our joint augmenting and contrasting framework, we generate stronger augmentation while keeping the hypergraph distribution with adversarial learning. This illustrates the importance of exploring the hypergraph structure. We also test our method on 1\% label setting in Table \ref{tab:1Results}. In this setting, Zoo and NTU2012 data sets are not shown because of the extremely small data size (each case has less than one training sample). We can find that in the 1\% label setting A4 (mask) method performs the best in  Cora, Citeseer and DBLP-CA. These data sets are all originally graphs, and are constructed as hypergraphs in different ways. So on these data sets, relatively little higher-order structural information can be explored with hypergraph structure perturbation-based contrastive learning.
% In the experiments, we have several important hyperparameters, we discuss how to set them in the appendix.

\begin{table}[t]
\setlength{\tabcolsep}{2.5pt}

\caption{Results on the test data sets: Mean accuracy (\%) $\pm$ standard deviation. Bold values indicate the best result. Underlined values indicate the second best. 10\% of all vertices are used for training.}
\label{tab:10Results}
\scriptsize
\centering
\scalebox{1.0}{
\begin{tabular}{ccccccccc}
\toprule
       & Cora                     & Citeseer                 & Pubmed                   & Cora-CA                  & DBLP-CA                  & Zoo                       & 20Newsgroups             & Mushroom                 \\ \midrule
SetGNN & 67.93 ± 1.27             & 63.53 ± 1.32             & 84.33 ± 0.36             & 72.21 ± 1.51             & 89.51 ± 0.18             & 65.06 ± 12.82             & 79.37 ± 0.35             & 99.75 ± 0.11             \\
Self   & 68.24 ± 1.12             & 62.49 ± 1.48             & 84.38 ± 0.38             & 72.74 ± 1.53             & 89.51 ± 0.23             & 57.35 ± 18.32             & 79.45 ± 0.32             & 95.83 ± 0.23             \\
Con    & 68.89 ± 1.80             & 62.82 ± 1.21             & 84.56 ± 0.34             & 73.22 ± 1.65             & 89.59 ± 0.13             & 61.05 ± 14.54             & 79.49 ± 0.45             & 95.85 ± 0.31             \\
A0     & 68.59 ± 1.33             & 62.25 ± 2.15             & 84.54 ± 0.42             & 71.85 ± 1.62             & 89.62 ± 0.24             & 62.57 ± 13.84             & 79.07 ± 0.46             & 99.77 ± 0.17             \\
A1     & 72.39 ± 1.34             & 66.28 ± 1.27             & 85.17 ± 0.37             & 75.45 ± 1.54             & 89.83 ± 0.21             & 65.80 ± 13.31             & 79.47 ± 0.32             & 99.80 ± 0.14             \\
A2     & 72.58 ± 1.09             & \underline{66.40 ± 1.35} & 85.16 ± 0.38             & \underline{75.62 ± 1.42} & \underline{90.22 ± 0.23} & \underline{66.35 ± 13.26} & \underline{79.56 ± 0.42} & 99.80 ± 0.17             \\
A3     & 72.33 ± 1.23             & 65.79 ± 1.18             & \underline{85.24 ± 0.28} & 75.34 ± 1.40             & 89.85 ± 0.16             & 65.79 ± 14.05             & 79.47 ± 0.34             & \underline{99.81 ± 0.10} \\
A4     & \underline{72.95 ± 1.19} & 66.22 ± 0.95             & 84.88 ± 0.38             & 75.29 ± 1.56             & 90.10 ± 0.18             & 62.59 ± 12.77             & 79.45 ± 0.48             & 99.80 ± 0.14             \\
A5     & 67.96 ± 0.99             & 63.21 ± 1.25             & 84.48 ± 0.40             & 72.61 ± 1.86             & 89.75 ± 0.24             & 62.47 ± 12.39             & 79.42 ± 0.52             & 99.79 ± 0.10             \\
A6     & \textbf{73.12 ± 1.48}    & \textbf{66.94 ± 1.00}    & \textbf{85.72 ± 0.38}    & \textbf{76.21 ± 1.26}    & \textbf{90.28 ± 0.19}    & \textbf{66.89 ± 12.44}    & \textbf{79.78 ± 0.40}    & \textbf{99.86 ± 0.10}    \\ \midrule
       & NTU2012                  & ModelNet40               & Yelp                     & House (0.6)              & House (1.0)              & Walmart (0.6)             & Walmart (1.0)            & Avg. Rank                \\ \midrule
SetGNN & 73.86 ± 1.62             & 95.85 ± 0.38             & 28.78 ± 1.51             & 68.54 ± 1.89             & 58.34 ± 2.25             & 74.97 ± 0.22              & 59.13 ± 0.20             & 7.71                     \\
Self   & 73.41 ± 1.65             & 95.83 ± 0.23             & 23.49 ± 4.15             & 67.75 ± 3.29             & 58.54 ± 2.16             & 74.76 ± 0.20              & 58.83 ± 0.21             & 8.64                     \\
Con    & 73.27 ± 1.53             & 95.85 ± 0.31             & 26.14 ± 1.86             & 68.50 ± 2.52             & 58.56 ± 2.42             & 75.17 ± 0.21              & 59.39 ± 0.20             & 7.07                     \\
A0     & 73.54 ± 1.93             & 95.92 ± 0.18             & 29.43 ± 1.42             & 67.48 ± 3.21             & 57.39 ± 2.37             & 73.14 ± 0.21              & 56.49 ± 0.60             & 8.21                     \\
A1     & 74.71 ± 1.81             & 95.87 ± 0.27             & 27.18 ± 0.71             & 68.64 ± 2.99             & 58.10 ± 3.22             & 75.42 ± 0.13              & 60.09 ± 0.25             & 4.50                     \\
A2     & \underline{74.88 ± 1.66} & \underline{96.56 ± 0.34} & \underline{31.39 ± 2.45} & \underline{69.73 ± 2.60} & 58.90 ± 1.97             & \underline{75.50 ± 0.18}  & \underline{60.19 ± 0.20} & \underline{2.29}         \\
A3     & 74.68 ± 1.74             & 96.48 ± 0.29             & 27.57 ± 1.00             & 67.88 ± 2.90             & 58.51 ± 2.22             & 75.29 ± 0.23              & 60.19 ± 0.20             & 4.71                     \\
A4     & 74.83 ± 1.75             & 95.86 ± 0.28             & 29.64 ± 1.93             & 69.56 ± 2.89             & \underline{58.91 ± 2.69} & 75.43 ± 0.18              & 59.90 ± 0.24             & 4.14                     \\
A5     & 74.41 ± 1.86             & 96.46 ± 0.33             & 29.24 ± 1.42             & 68.14 ± 2.97             & 57.70 ± 2.98             & 75.26 ± 0.18              & 59.81 ± 0.22             & 6.71                     \\
A6     & \textbf{75.34 ± 1.91}    & \textbf{96.93 ± 0.33}    & \textbf{34.64 ± 0.39}    & \textbf{70.96 ± 2.27}    & \textbf{59.93 ± 1.99}    & \textbf{75.62 ± 0.16}     & \textbf{60.46 ± 0.20}    & \textbf{1.00}            \\ \bottomrule
\end{tabular}}
% \vspace{-0.6cm}
\end{table}

\begin{table}[t]
\setlength{\tabcolsep}{2.5pt}

\caption{Results on the test data sets: Mean accuracy (\%) $\pm$ standard deviation. Bold values indicate the best result. 1\% of all vertexes are used for training.}
\label{tab:1Results}
\scriptsize
\centering
\scalebox{1.0}{
\begin{tabular}{cccccccc}
\toprule
       & Cora                     & Citeseer                 & Pubmed                   & Cora-CA                  & DBLP-CA                  & 20Newsgroups             & Mushroom                 \\ \midrule
SetGNN & 46.48 ± 3.62             & 47.01 ± 4.31             & 76.13 ± 1.19             & 52.29 ± 4.18             & 85.52 ± 0.54             & 73.83 ± 1.40             & 97.73 ± 1.18             \\
Self   & 45.79 ± 5.34             & 44.22 ± 4.43             & 76.71 ± 0.90             & 51.64 ± 5.37             & 84.42 ± 0.37             & 73.91 ± 0.90             & 92.25 ± 0.89             \\
Con    & 49.20 ± 4.38             & 48.56 ± 4.88             & 77.51 ± 1.08             & 52.37 ± 4.41             & 86.47 ± 0.35             & 74.39 ± 1.23             & 92.43 ± 0.87             \\
A0     & 48.50 ± 4.77             & 46.43 ± 4.24             & 78.83 ± 1.79             & 49.87 ± 5.08             & 87.34 ± 0.73             & 74.43 ± 1.11             & 97.32 ± 1.33             \\
A1     & 56.42 ± 5.02             & 55.63 ± 3.96             & 80.13 ± 1.44             & 60.86 ± 5.91             & 87.53 ± 0.30             & 74.68 ± 1.31             & 97.95 ± 1.15             \\
A2     & 56.81 ± 4.49             & 56.10 ± 2.86             & 80.22 ± 1.24             & \underline{60.96 ± 6.31} & 88.10 ± 0.35             & 74.72 ± 1.16             & \underline{98.05 ± 1.18} \\
A3     & 55.94 ± 3.67             & 55.82 ± 3.40             & 80.13 ± 1.02             & 60.51 ± 4.55             & 87.47 ± 0.36             & 74.63 ± 1.00             & 98.04 ± 0.98             \\
A4     & \textbf{58.55 ± 5.14}    & \textbf{57.16 ± 4.62}    & \underline{80.11 ± 1.02} & 60.91 ± 5.15             & \textbf{88.91 ± 0.29}    & 74.67 ± 1.39             & 97.72 ± 1.12             \\
A5     & 46.23 ± 3.44             & 45.07 ± 4.89             & 75.95 ± 1.32             & 53.26 ± 4.86             & 87.12 ± 0.43             & \underline{74.81 ± 1.04} & 97.72 ± 1.25             \\
A6     & \underline{57.45 ± 5.00} & \underline{56.23 ± 3.27} & \textbf{81.10 ± 0.80}    & \textbf{61.76 ± 4.94}    & \underline{88.55 ± 0.41} & \textbf{75.52 ± 0.93}    & \textbf{98.28 ± 1.03}    \\ \midrule
       & ModelNet40               & Yelp                     & House (0.6)              & House (1.0)              & Walmart (0.6)            & Walmart (1.0)            & Avg. Rank ($\downarrow$)                \\ \midrule
SetGNN & 88.34 ± 2.69             & 27.64 ± 1.10             & 53.69 ± 2.20             & 51.85 ± 1.64             & 65.48 ± 0.45             & 51.15 ± 0.52             & 7.62                     \\
Self   & 86.85 ± 3.03             & 20.77 ± 5.15             & 53.42 ± 2.25             & 51.14 ± 1.75             & 65.23 ± 0.43             & 51.00 ± 0.41             & 9.69                     \\
Con    & 87.00 ± 2.99             & 24.23 ± 0.43             & 53.58 ± 3.04             & 51.96 ± 1.87             & 65.47 ± 0.44             & 51.13 ± 0.46             & 7.31                     \\
A0     & 88.75 ± 2.78             & 27.43 ± 0.60             & 53.60 ± 2.73             & 51.70 ± 2.13             & 65.41 ± 0.47             & 51.10 ± 0.49             & 7.46                     \\
A1     & 89.34 ± 2.66             & 26.18 ± 0.51             & 54.12 ± 3.29             & 52.23 ± 2.46             & 65.96 ± 0.36             & 51.22 ± 0.35             & 4.08                     \\
A2     & 89.37 ± 2.69             & 27.67 ± 0.91             & \underline{54.42 ± 2.83} & \underline{52.31 ± 1.44} & \underline{66.01 ± 0.41} & \underline{51.32 ± 0.30} & \underline{2.69}         \\
A3     & 89.31 ± 2.62             & 26.98 ± 0.66             & 53.71 ± 2.71             & 52.11 ± 2.24             & 65.88 ± 0.50             & 51.35 ± 0.53             & 4.38                     \\
A4     & 89.03 ± 2.66             & 27.45 ± 0.81             & 53.64 ± 2.61             & 51.77 ± 2.20             & 65.55 ± 0.51             & 51.04 ± 0.47             & 4.54                     \\
A5     & \underline{89.43 ± 2.68} & \underline{28.09 ± 0.96} & 54.07 ± 3.09             & 51.94 ± 1.84             & 65.52 ± 0.39             & 50.97 ± 0.47             & 6.00                     \\
A6     & \textbf{90.22 ± 2.72}    & \textbf{29.61 ± 0.71}    & \textbf{56.27 ± 4.18}    & \textbf{52.55 ± 2.18}    & \textbf{66.42 ± 0.40}    & \textbf{51.82 ± 0.39}    & \textbf{1.23}            \\ \bottomrule
\end{tabular}}
% \vspace{-0.3cm}
\end{table}

\begin{table}[t]
\setlength{\tabcolsep}{2.5pt}
\caption{Results of different self-supervised mechanisms: Mean accuracy (\%) $\pm$ standard deviation. Bold values indicate the best result. 10\% of all vertexes are used for training.}
\label{tab:mechanism}
\scriptsize
\centering
\scalebox{1.0}{
\begin{tabular}{cccccccc}
\toprule
                      & Cora                  & Citeseer              & Pubmed                & Cora-CA               & DBLP-CA               & Zoo                    & 20Newsgroups          \\ \midrule
SetGNN                & 67.93 ± 1.27          & 63.53 ± 1.32          & 84.33 ± 0.36          & 72.21 ± 1.51          & 89.51 ± 0.18          & 65.06 ± 12.82          & 79.37 ± 0.35          \\
Pretrain\_L           & 52.59 ± 2.33          & 53.29 ± 2.01          & 69.90 ± 0.41          & 48.00 ± 4.79          & 87.59 ± 0.43          & \textbf{66.82 ± 13.48} & 71.93 ± 2.99          \\
Pretrain\_F           & \underline{68.39 ± 1.20}          & \underline{63.83 ± 1.68}          & \underline{84.47 ± 0.40}          & \underline{73.12 ± 1.37}          & \underline{89.75 ± 0.23}          & 65.43 ± 13.38          & \underline{79.44 ± 0.39}          \\
MTL                   & \textbf{72.58 ± 1.10} & \textbf{66.40 ± 1.35} & \textbf{85.16 ± 0.38} & \textbf{75.82 ± 1.42} & \textbf{90.22 ± 0.23} & \underline{66.35 ± 13.26}          & \textbf{79.56 ± 0.42} \\ \midrule
Mushroom              & NTU2012               & ModelNet40            & Yelp                  & House (0.6)           & House (1.0)           & Walmart (0.6)          & Walmart (1.0)         \\ \midrule
\underline{99.75 ± 0.11}          & 73.86 ± 1.62          & 95.85 ± 0.38          & \underline{28.78 ± 1.51}          & 68.54 ± 1.89          & 58.34 ± 2.25          & 74.97 ± 0.22           & 59.13 ± 0.20          \\
93.77 ± 2.20          & 70.06 ± 2.42          & \underline{96.23 ± 0.31}          & 26.68 ± 0.30          & 61.22 ± 3.09          & 54.81 ± 2.39          & 40.35 ± 4.30           & 33.30 ± 2.72          \\
99.77 ± 0.15          & \underline{74.03 ± 1.86}          & 95.88 ± 0.34          & 28.19 ± 1.42          & \underline{69.02 ± 4.02}          & \textbf{59.20 ± 2.54} & \underline{75.01 ± 0.27}           & \underline{59.87 ± 0.28}          \\
\textbf{99.80 ± 0.17} & \textbf{74.88 ± 1.66} & \textbf{96.56 ± 0.34} & \textbf{31.39 ± 2.45} & \textbf{69.73 ± 2.60} & \underline{58.90 ± 1.97}          & \textbf{75.50 ± 0.18}  & \textbf{60.19 ± 0.20} \\ \bottomrule
\end{tabular}}
% \vspace{-0.5cm}
\end{table}

\begin{table}[t]
\setlength{\tabcolsep}{2.5pt}

\caption{Results on converted conventional graphs: Mean accuracy (\%) $\pm$ standard deviation. Bold values indicate the best result. 10\% of all vertices are used for training.}
\label{tab:cgraph}
\scriptsize
\centering
\scalebox{1.0}{
\begin{tabular}{cccccccc}
\midrule
                      & Cora                  & Citeseer              & Pubmed                & Cora-CA               & DBLP-CA               & Zoo                   & 20Newsgroups          \\ \toprule
HGNN                  & \textbf{67.37 ± 1.45} & 62.76 ± 1.42          & 82.16 ± 0.38          & 66.80 ± 1.79          & \textbf{85.28 ± 0.29} & \textbf{47.84 ± 6.87} & \underline{70.27 ± 0.73}          \\
A2                    & 67.18 ± 1.42          & \textbf{63.52 ± 2.35} & \textbf{82.37 ± 0.34} & \textbf{67.14 ± 1.79} & \underline{85.22 ± 0.26}          & 46.85 ± 10.15         & \textbf{70.46 ± 1.21} \\
A4                    & \underline{67.24 ± 1.51}          & \underline{63.37 ± 2.56}          & \underline{82.25 ± 0.43}          & \underline{66.88 ± 2.07}          & 85.16 ± 0.25          & \underline{46.85 ± 9.93}          & 69.35 ± 1.24          \\ \midrule
Mushroom              & NTU2012               & ModelNet40            & Yelp                  & House (0.6)           & House (1.0)           & Walmart (0.6)         & Walmart (1.0)         \\ \midrule
\underline{97.15 ± 0.47}          & \textbf{70.26 ± 1.70} & 87.60 ± 0.36          & \textbf{26.91 ± 0.37} & 58.01 ± 2.47          & \underline{57.65 ± 2.69}          & 59.48 ± 0.19          & 53.97 ± 0.29          \\
\textbf{97.29 ± 0.45} & 69.91 ± 1.59          & \textbf{87.75 ± 0.33} & \underline{26.72 ± 0.36}          & \underline{58.08 ± 3.28}          & 57.53 ± 2.80          & \underline{59.49 ± 0.22}          & \textbf{54.04 ± 0.24} \\
97.15 ± 0.55          & \underline{69.94 ± 1.54}          & \underline{87.65 ± 0.36}          & 26.66 ± 0.45          & \textbf{58.47 ± 2.99} & \textbf{57.73 ± 2.84} & \textbf{59.53 ± 0.22} & \underline{53.98 ± 0.26}          \\ \bottomrule
\end{tabular}}
% \vspace{-0.3cm}
\end{table}

\textbf{Comparison between multi-task learning and pretraining.}
We then compare the multi-task training method with the pretraining method in Table \ref{tab:mechanism}. Pretrain\_L adopts the linear evaluation protocol where a linear classifier is trained on top of the fixed pretrained representations. Pretrain\_F follows a fully finetuning protocol that uses the weights of the learned hypergnn encoder as initialization while finetuning all the layers. MTL denotes the multi-task learning method which trains the supervised classification loss and contrastive loss together. For all the methods, we use A2 (generalized hyperedge perturbation) as it performs the best among fabricated augmentations. From the table, we find MTL achieves the best performance in nearly all data sets. Pretrain\_L and Pretrain\_F can only obtain better performance on two small data sets: Zoo and House. We find on most data sets, Pretrain\_L makes the model perform worse, which shows that the linear classifier is not enough to represent the higher-order information in the hypergraph. Pretrain\_F has a much better performance compared with Pretrain\_L, which indicates the effects of using contrastive learning. However, the method switches the objective during finetuning, which would lead to the memorization problem and the loss of pre-trained knowledge. Therefore, all our other experiments adopt MTL setting to incorporate contrastive self-supervision.

\textbf{Comparison to the converted graph.}
Next we investigate on which graph should we contrast on. We first convert the original hypergraph into a conventional graph using the clique expansion technique, and we choose the representative HGNN\cite{feng2019hypergraph} as the backbone network for learning on the converted graph. We compare it with two representative augmentations:  A2 (Here, edge perturbation on the conventional graph) and A4 (feature masking). In Table \ref{tab:cgraph}, we find that HGNN performs poorly compared with SetGNN. Apparently contrastive self-supervision on converted graphs does not bring much benefit. The reason is that part of structural information, important to hypergraph representation learning, is lost when hypergraphs are converted to graphs.  
%during the conversion process. Then we can see that the contrastive self-supervision seldom brings benefits to the model due to the fact that it is unable to fully explore the rich structural information of hypergraphs. 
These results  indicate the importance of designing HyperGNN and contrastive strategies directly on hypergraphs.

\begin{figure}[t] 
\centering
\includegraphics[width=0.9\linewidth]{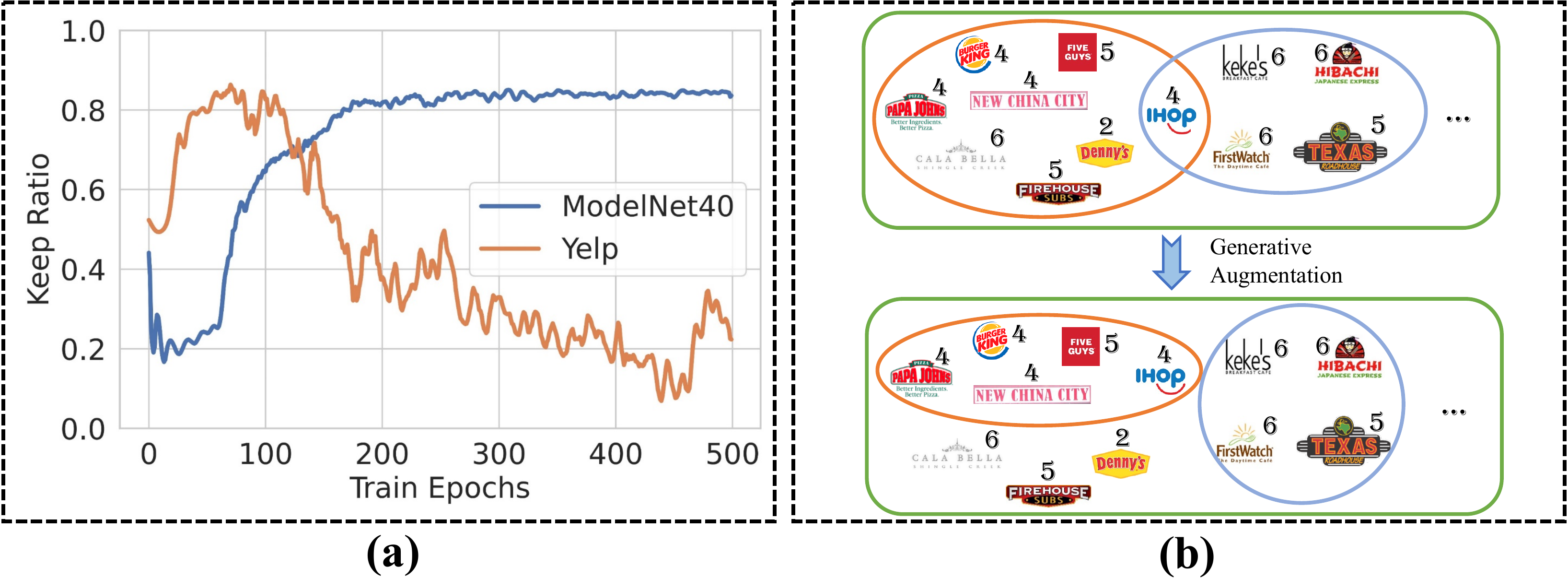}
\caption{(a) Training dynamics of the relation keep ratio. (b) Illustration of our proposed generative augmentation on the Yelp data set. Each icon represents a restaurant in the data set, and the number near the icon is the label of this restaurant. The ellipse denotes the hyperedge.}
% \vspace{-0.3cm}
\label{fig:vis}
% \vspace{-0.7cm}
\end{figure}

\textbf{Analysis of generative augmentation.}
Here we analyze our proposed generative hypergraph augmentation (A6). We select ModelNet40 and Yelp as the representatives of high-homophily and low-homophily data sets, respectively. 

First, we examine the training dynamics of keep ratio and find that they are  highly related to the data set homophily (Figure \ref{fig:vis} (a)). For ModelNet40, the generator keeps only 20\% relations in the early training stage. This is because the homophily of the data set is very high (0.92/0.88) and deleting a lot of edges relatively randomly at the beginning would not have a large impact on the model but rather can help model learn structural information. Then at a later stage, the generator begins to keep more than 80\% relations, which means that the model has learned higher-order information and only removes the unnecessary relations. For Yelp, a similar conclusion holds. Specifically, as its homophily is pretty low (0.57/0.26),  the generator keeps most of the relations at the early stage for training and then just keeps a very low ratio of related relations at the later training stage. 

Next, we investigate what our generator has  learned. We visualize the hyperedges in the Yelp data set in Figure \ref{fig:vis} (b). Yelp is a restaurant-rating data set and the restaurants visited by the same user are connected by a hyperedge. We find that some vertices with different labels are removed, which could remove extraneous information and improve the hypergraph homophily.   
%This removes extraneous information and improves the homophily of the hypergraph. 
This indicates that our generator does grasp the higher-order information in the hypergraph. %YS:???

% \begin{figure}[t] 
% \centering
% \subfigure{
% \begin{minipage}[t]{0.45\textwidth}
% \centering
% % \RaggedRight
% \includegraphics[scale=0.35]{image/ratio.jpg}
% \label{fig:ratio}
% \end{minipage}
% }
% \centering
% \subfigure{
% \begin{minipage}[t]{0.4\textwidth}
% \centering
% % \RaggedRight
% \includegraphics[scale=0.25]{image/vis.pdf}
% \label{fig:ratio}
% \end{minipage}
% }
% \caption{(a). (b). }
% % \vspace{-0.3cm}
% \end{figure}

\begin{table}[t]
\setlength{\tabcolsep}{2.5pt}

\caption{Results on the test data sets with regard to robustness. Bold values indicate the best result. 10\% of all vertexes are used for training.}
\label{tab:robust}
\scriptsize
\centering
\scalebox{1.0}{
\begin{tabular}{cccccccccc}
\hline
\multicolumn{1}{c|}{} & \multicolumn{3}{c|}{Cora}                                                      & \multicolumn{3}{c|}{Citeseer}                                                  & \multicolumn{3}{c}{ModelNet40}                                                 \\ \hline
                      & Random                   & Net                      & Minmax                   & Random                   & Net                      & Minmax                   & Random                   & Net                      & Minmax                   \\ \hline
SetGNN                & 66.87 ± 1.33             & 66.26 ± 1.54             & 66.58 ± 1.02             & 62.89 ± 1.57             & 62.81 ± 1.32             & 62.21 ± 1.64             & 95.74 ± 0.22             & 95.41 ± 0.28             & 93.33 ± 0.26             \\
A2                    & 71.90 ± 1.63             & \underline{71.16 ± 0.92} & \underline{70.86 ± 1.22} & \underline{66.41 ± 1.08} & \underline{65.38 ± 1.47} & \underline{64.69 ± 0.98} & \underline{96.09 ± 0.17} & \underline{95.52 ± 0.24} & \underline{93.64 ± 0.26} \\
A4                    & \underline{72.11 ± 1.60} & 70.49 ± 1.29             & 70.52 ± 1.39             & 65.94 ± 1.24             & 65.15 ± 1.70             & 64.12 ± 1.19             & 95.79 ± 0.27             & 95.44 ± 0.25             & 93.35 ± 0.24             \\
A6                    & \textbf{72.15 ± 1.70}    & \textbf{71.94 ± 1.48}    & \textbf{71.98 ± 1.36}    & \textbf{66.60 ± 1.61}    & \textbf{65.68 ± 1.09}    & \textbf{65.51 ± 1.13}    & \textbf{96.58 ± 0.24}    & \textbf{96.23 ± 0.23}    & \textbf{94.82 ± 0.33}    \\ \hline
\multicolumn{1}{c|}{} & \multicolumn{3}{c|}{NTU2012}                                                   & \multicolumn{3}{c|}{House (0.6)}                                               & \multicolumn{3}{c}{House (1.0)}                                                \\ \hline
                      & Random                   & Net                      & Minmax                   & Random                   & Net                      & Minmax                   & Random                   & Net                      & Minmax                   \\ \hline
SetGNN                & 73.84 ± 2.18             & 73.38 ± 1.36             & 70.71 ± 1.89             & 67.16 ± 2.55             & 68.88 ± 2.68             & 64.78 ± 2.20             & 56.86 ± 1.93             & 59.95 ± 1.92             & 56.52 ± 2.52             \\
A2                    & \underline{74.50 ± 2.03} & \underline{73.86 ± 1.84} & \underline{71.40 ± 1.64} & \underline{67.71 ± 2.94} & \underline{69.59 ± 2.32} & \underline{65.23 ± 2.89} & \underline{57.74 ± 2.70} & \underline{60.73 ± 2.30} & \underline{57.00 ± 1.94} \\
A4                    & 73.73 ± 1.59             & 73.72 ± 1.59             & 71.06 ± 1.53             & 67.55 ± 2.41             & 68.85 ± 1.38             & 64.97 ± 3.35             & 57.47 ± 2.72             & 60.10 ± 1.74             & 56.65 ± 2.26             \\
A6                    & \textbf{75.06 ± 1.97}    & \textbf{74.37 ± 1.99}    & \textbf{72.09 ± 1.98}    & \textbf{69.88 ± 3.27}    & \textbf{73.14 ± 2.71}    & \textbf{68.84 ± 2.71}    & \textbf{60.06 ± 2.07}    & \textbf{62.41 ± 1.77}    & \textbf{58.76 ± 2.24}    \\ \hline
\end{tabular}}
% \vspace{-0.3cm}
\end{table}

\begin{table}[t]
\setlength{\tabcolsep}{2.5pt}

\caption{Results on the test data sets with regard to fairness. 10\% of all vertexes are used for training. For fairness metrics $\Delta_{SP}$ and $\Delta_{EO}$, lower values indicate better performance.}
\label{tab:fair}
\small
\centering
\scalebox{0.8}{
\begin{tabular}{cccccc}
\toprule
data set                           & Method & AUROC                 & F1                    & $\Delta_{SP} (\downarrow)$        & $\Delta_{EO} (\downarrow)$        \\ \midrule
\multirow{4}{*}{German Credit}    & SetGNN & 59.16 ± 2.51          & 81.84 ± 0.93          & 2.65 ± 5.62          & 4.06 ± 6.76          \\
                                & A2   & \underline{59.81 ± 3.00}          & \underline{82.26 ± 0.13}          & \textbf{0.55 ± 0.95} & \underline{0.78 ± 0.70}          \\
                                  & A4   & 59.66 ± 3.83          & 80.54 ± 3.52          & 3.03 ± 6.54          & 5.07 ± 7.81          \\
                                  & A6    & \textbf{59.88 ± 3.04} & \textbf{82.36 ± 0.38} & \underline{0.95 ± 0.92}          & \textbf{0.47 ± 0.56} \\ \midrule
\multirow{4}{*}{Recidivism}       & SetGNN & \underline{96.51 ± 0.48}          & \underline{89.84 ± 0.97}          & 8.63 ± 0.50          & 4.16 ± 0.51          \\
                                    & A2   & 96.34 ± 0.39          & \textbf{90.09 ± 0.53} & 8.53 ± 0.52          & 3.92 ± 0.68          \\
                                  & A4   & 96.45 ± 0.35          & 89.75 ± 0.68          & \textbf{8.49 ± 0.27} & \underline{3.49 ± 0.66}          \\
                                  
                                  & A6    & \textbf{96.55 ± 0.54} & 89.22 ± 0.55          & \underline{8.51 ± 0.25}          & \textbf{3.13 ± 0.64} \\ \midrule
\multirow{4}{*}{Credit defaulter} & SetGNN & 73.46 ± 0.17          & 87.91 ± 0.27          & 2.79 ± 0.99          & 0.98 ± 0.69          \\
                                & A2   & 73.43 ± 0.27          & 87.82 ± 0.24          & \underline{2.64 ± 1.32}          & \underline{0.93 ± 0.87}          \\
                                  & A4   & \underline{73.58 ± 0.19}          & \underline{87.92 ± 0.25}          & 2.84 ± 1.14          & 1.38 ± 0.32          \\
                                  
                                  & A6    & \textbf{73.78 ± 0.16} & \textbf{88.03 ± 0.14} & \textbf{2.58 ± 0.91} & \textbf{0.81 ± 0.37} \\ \bottomrule
\end{tabular}}
% \vspace{-0.6cm}
\end{table}

\textbf{Adversarial robustness.}
Besides generalizability, we here show hypergraph contrastive learning also boosts robustness.
Since there is no existing work developing adversarial attack algorithms designated for hypergraphs, we adapt two state-of-the-art attackers from the graph domain. We presume the graph attackers are applicable to hypergraphs to a certain extent, both of which are non-Euclidean data structures with sharing properties. The experiments are performed on five real-world data sets: Cora, Citeseer, ModelNet40, NTU2012 and House.
% Since the robustness of hypergraphs has not been explored before, we first introduce two adapted attack algorithms from the conventional graph to evaluate the model because we think graph and hypergraph share some common properties.
We regard each hypergraph as a bipartite graph and leverage those algorithms to conduct attacks to the vertices and hyperedges. The methods include an untargeted attack method, minmax attack \cite{zugner2019adversarial} which poisons the graph structures by adding and removing relations to reduce the overall performance; and a targeted attack method, nettack \cite{zugner2018adversarial} which leads the HyperGNN to mis-classify target vertices. Beyond these, we also include a random hypergraph perturbation baseline which will randomly drop numerous relations in the hypergraph. These three methods are denoted as Net, Minmax and Random. Following previous works, for each attack, we perturb 10\% of vertices/relations. The results in Table \ref{tab:robust} show that Random and Minmax attacks can decrease the performance of the original model a little on all the data sets, while net attack can decrease the performance on most data sets and surprisingly increase the performance on the House data set. This performance gain indicates HyperGNN is more robust to structure attack compared with GNN as it leverages higher-order information. Based on these, HyperGCL with generalized hyperedge augmentation (A2) performs better than feature perturbation (A4), and our proposed generative augmentation (A6) can surpass these fabricated baselines on all the data sets and thus is the best to defend attacks. We believe this’ll be a beneficial complement to our main experiments and we hope for more works on hypergraph attacks.

\textbf{Fairness.}
Furthermore, we claim that hypergraph contrastive self-supervision also benefits fairness. There was no related data set before. So we introduce three newly curated hypergraph data sets: German \cite{asuncion2007uci}, Recidivism \cite{jordan2015effect} and Credit \cite{yeh2009comparisons}. The hypergraph construction follows the setting in \cite{feng2019hypergraph}. The top 5 similar objects in each data set are built as a hyperedge. For the accuracy metrics, we use F1-score and AUROC value for the binary classification task. For measuring fairness, we adopt the statistical parity $\Delta_{SP}$ and equalized odds $\Delta_{EO}$. Please refer to Appendix C for detailed information about the data sets and metrics. The experimental results in Table \ref{tab:fair} show that our generative method still achieves better or comparable performances while imposing more fairness.

%The German data set \cite{asuncion2007uci} is also from the machine learning category and has 1,000 clients in a German bank. Gender is regarded as the sensitive attribute. The Recidivism
% data set has 18,876 defendants who got released on bail at the U.S state courts during 1990-2009.

% \subsection{Hyperparameter Analysis}
% loss trade-off parameter, dropout ratio, augmentation ratio.
% Properly set the dropout ratio can improve performance, and oversetting the dropout ratio can cause cl to crash.

% \vspace{-1em}
\section{Conclusion}
\label{section:conclusion}
% \vspace{-1em}
In the paper, we study the problem of how to construct contrastive views of hypergraphs via augmentations. We provide the solutions by first studying domain knowledge-guided fabrication schemes. Then, in search of more effective views in a data-driven manner, we are the first to propose hypergraph generative models to generate augmented views, as well as an end-to-end differentiable pipeline to jointly perform hypergraph augmentation and contrastive learning. We find that generative augmentations perform better at preserving higher-order information to further benefit generalizability. The proposed framework also boosts robustness and fairness of hypergraph representation learning. In the future, we plan to design more powerful hypergraph generator and HyperGNN while addressing more real-world hypergraph data challenges and more hypergraph learning models.

\begin{ack}
This work is supported by National Science Foundation under Award No. IIS-1947203, IIS-2117902, IIS-2137468, CCF-1943008; US Army Research Office
Young Investigator Award W911NF2010240; National Institute of General Medical Sciences under grant R35GM124952; and Agriculture and Food Research Initiative grant no. 2020-67021-32799/project accession no.1024178 from the USDA National Institute of Food and Agriculture. The views and conclusions are those of the authors and should not be interpreted as representing the official policies of the government agencies.
\end{ack}

% {\small
\bibliographystyle{unsrt}
\bibliography{ref}

\begin{thebibliography}{10}

\bibitem{feng2019hypergraph}
Yifan Feng, Haoxuan You, Zizhao Zhang, Rongrong Ji, and Yue Gao.
\newblock Hypergraph neural networks.
\newblock In {\em Proceedings of the AAAI Conference on Artificial
  Intelligence}, volume~33, pages 3558--3565, 2019.

\bibitem{yadati2019hypergcn}
Naganand Yadati, Madhav Nimishakavi, Prateek Yadav, Vikram Nitin, Anand Louis,
  and Partha Talukdar.
\newblock Hypergcn: A new method for training graph convolutional networks on
  hypergraphs.
\newblock {\em Advances in neural information processing systems}, 32, 2019.

\bibitem{chien2021you}
Eli Chien, Chao Pan, Jianhao Peng, and Olgica Milenkovic.
\newblock You are allset: A multiset function framework for hypergraph neural
  networks.
\newblock {\em ICLR}, 2022.

\bibitem{bretto2013hypergraph}
Alain Bretto.
\newblock Hypergraph theory.
\newblock {\em An introduction. Mathematical Engineering. Cham: Springer},
  2013.

\bibitem{ji2020dual}
Shuyi Ji, Yifan Feng, Rongrong Ji, Xibin Zhao, Wanwan Tang, and Yue Gao.
\newblock Dual channel hypergraph collaborative filtering.
\newblock In {\em Proceedings of the 26th ACM SIGKDD International Conference
  on Knowledge Discovery \& Data Mining}, pages 2020--2029, 2020.

\bibitem{yu2021self}
Junliang Yu, Hongzhi Yin, Jundong Li, Qinyong Wang, Nguyen Quoc~Viet Hung, and
  Xiangliang Zhang.
\newblock Self-supervised multi-channel hypergraph convolutional network for
  social recommendation.
\newblock In {\em Proceedings of the Web Conference 2021}, pages 413--424,
  2021.

\bibitem{sawhney2020spatiotemporal}
Ramit Sawhney, Shivam Agarwal, Arnav Wadhwa, and Rajiv~Ratn Shah.
\newblock Spatiotemporal hypergraph convolution network for stock movement
  forecasting.
\newblock In {\em 2020 IEEE International Conference on Data Mining (ICDM)},
  pages 482--491. IEEE, 2020.

\bibitem{you2022cross}
Yuning You and Yang Shen.
\newblock Cross-modality and self-supervised protein embedding for
  compound-protein affinity and contact prediction.
\newblock {\em bioRxiv}, 2022.

\bibitem{zhang2022multiscale}
Ruochi Zhang, Tianming Zhou, and Jian Ma.
\newblock Multiscale and integrative single-cell hi-c analysis with higashi.
\newblock {\em Nature biotechnology}, 40(2):254--261, 2022.

\bibitem{liu2022generating}
Meng Liu, Youzhi Luo, Kanji Uchino, Koji Maruhashi, and Shuiwang Ji.
\newblock Generating 3d molecules for target protein binding.
\newblock {\em arXiv preprint arXiv:2204.09410}, 2022.

\bibitem{goyal2019scaling}
Priya Goyal, Dhruv Mahajan, Abhinav Gupta, and Ishan Misra.
\newblock Scaling and benchmarking self-supervised visual representation
  learning.
\newblock In {\em Proceedings of the ieee/cvf International Conference on
  computer vision}, pages 6391--6400, 2019.

\bibitem{chen2020simple}
Ting Chen, Simon Kornblith, Mohammad Norouzi, and Geoffrey Hinton.
\newblock A simple framework for contrastive learning of visual
  representations.
\newblock In {\em International conference on machine learning}, pages
  1597--1607. PMLR, 2020.

\bibitem{hu2019strategies}
Weihua Hu, Bowen Liu, Joseph Gomes, Marinka Zitnik, Percy Liang, Vijay Pande,
  and Jure Leskovec.
\newblock Strategies for pre-training graph neural networks.
\newblock {\em arXiv preprint arXiv:1905.12265}, 2019.

\bibitem{you2020graph}
Yuning You, Tianlong Chen, Yongduo Sui, Ting Chen, Zhangyang Wang, and Yang
  Shen.
\newblock Graph contrastive learning with augmentations.
\newblock {\em Advances in Neural Information Processing Systems},
  33:5812--5823, 2020.

\bibitem{you2020does}
Yuning You, Tianlong Chen, Zhangyang Wang, and Yang Shen.
\newblock When does self-supervision help graph convolutional networks?
\newblock {\em arXiv preprint arXiv:2006.09136}, 2020.

\bibitem{jin2020self}
Wei Jin, Tyler Derr, Haochen Liu, Yiqi Wang, Suhang Wang, Zitao Liu, and
  Jiliang Tang.
\newblock Self-supervised learning on graphs: Deep insights and new direction.
\newblock {\em arXiv preprint arXiv:2006.10141}, 2020.

\bibitem{xie2022self}
Yaochen Xie, Zhao Xu, Jingtun Zhang, Zhengyang Wang, and Shuiwang Ji.
\newblock Self-supervised learning of graph neural networks: A unified review.
\newblock {\em IEEE Transactions on Pattern Analysis and Machine Intelligence},
  2022.

\bibitem{velickovic2019deep}
Petar Velickovic, William Fedus, William~L Hamilton, Pietro Li{\`o}, Yoshua
  Bengio, and R~Devon Hjelm.
\newblock Deep graph infomax.
\newblock {\em ICLR (Poster)}, 2(3):4, 2019.

\bibitem{you2021graph}
Yuning You, Tianlong Chen, Yang Shen, and Zhangyang Wang.
\newblock Graph contrastive learning automated.
\newblock In {\em International Conference on Machine Learning}, pages
  12121--12132. PMLR, 2021.

\bibitem{you2022bringing}
Yuning You, Tianlong Chen, Zhangyang Wang, and Yang Shen.
\newblock Bringing your own view: Graph contrastive learning without
  prefabricated data augmentations.
\newblock {\em arXiv preprint arXiv:2201.01702}, 2022.

\bibitem{zhu2020deep}
Yanqiao Zhu, Yichen Xu, Feng Yu, Qiang Liu, Shu Wu, and Liang Wang.
\newblock Deep graph contrastive representation learning.
\newblock {\em arXiv preprint arXiv:2006.04131}, 2020.

\bibitem{trivedi2022augmentations}
Puja Trivedi, Ekdeep~Singh Lubana, Yujun Yan, Yaoqing Yang, and Danai Koutra.
\newblock Augmentations in graph contrastive learning: Current methodological
  flaws \& towards better practices.
\newblock In {\em Proceedings of the ACM Web Conference 2022}, pages
  1538--1549, 2022.

\bibitem{wang2022augmentation}
Haonan Wang, Jieyu Zhang, Qi~Zhu, and Wei Huang.
\newblock Augmentation-free graph contrastive learning.
\newblock {\em arXiv preprint arXiv:2204.04874}, 2022.

\bibitem{feng2022adversarial}
Shengyu Feng, Baoyu Jing, Yada Zhu, and Hanghang Tong.
\newblock Adversarial graph contrastive learning with information
  regularization.
\newblock In {\em Proceedings of the ACM Web Conference 2022}, pages
  1362--1371, 2022.

\bibitem{xu2022group}
Xinyi Xu, Cheng Deng, Yaochen Xie, and Shuiwang Ji.
\newblock Group contrastive self-supervised learning on graphs.
\newblock {\em IEEE Transactions on Pattern Analysis and Machine Intelligence},
  2022.

\bibitem{xia2022hypergraph}
Lianghao Xia, Chao Huang, Yong Xu, Jiashu Zhao, Dawei Yin, and Jimmy~Xiangji
  Huang.
\newblock Hypergraph contrastive collaborative filtering.
\newblock {\em arXiv preprint arXiv:2204.12200}, 2022.

\bibitem{cai2022hypergraph}
Derun Cai, Chenxi Sun, Moxian Song, Baofeng Zhang, Shenda Hong, and Hongyan Li.
\newblock Hypergraph contrastive learning for electronic health records.
\newblock In {\em Proceedings of the 2022 SIAM International Conference on Data
  Mining (SDM)}, pages 127--135. SIAM, 2022.

\bibitem{jang2016categorical}
Eric Jang, Shixiang Gu, and Ben Poole.
\newblock Categorical reparameterization with gumbel-softmax.
\newblock {\em arXiv preprint arXiv:1611.01144}, 2016.

\bibitem{zhang2019hyper}
Ruochi Zhang, Yuesong Zou, and Jian Ma.
\newblock Hyper-sagnn: a self-attention based graph neural network for
  hypergraphs.
\newblock {\em arXiv preprint arXiv:1911.02613}, 2019.

\bibitem{bai2021hypergraph}
Song Bai, Feihu Zhang, and Philip~HS Torr.
\newblock Hypergraph convolution and hypergraph attention.
\newblock {\em Pattern Recognition}, 110:107637, 2021.

\bibitem{huang2021unignn}
Jing Huang and Jie Yang.
\newblock Unignn: a unified framework for graph and hypergraph neural networks.
\newblock {\em arXiv preprint arXiv:2105.00956}, 2021.

\bibitem{arya2020hypersage}
Devanshu Arya, Deepak~K Gupta, Stevan Rudinac, and Marcel Worring.
\newblock Hypersage: Generalizing inductive representation learning on
  hypergraphs.
\newblock {\em arXiv preprint arXiv:2010.04558}, 2020.

\bibitem{zaheer2017deep}
Manzil Zaheer, Satwik Kottur, Siamak Ravanbakhsh, Barnabas Poczos, Russ~R
  Salakhutdinov, and Alexander~J Smola.
\newblock Deep sets.
\newblock {\em Advances in neural information processing systems}, 30, 2017.

\bibitem{he2020momentum}
Kaiming He, Haoqi Fan, Yuxin Wu, Saining Xie, and Ross Girshick.
\newblock Momentum contrast for unsupervised visual representation learning.
\newblock In {\em Proceedings of the IEEE/CVF conference on computer vision and
  pattern recognition}, pages 9729--9738, 2020.

\bibitem{grill2020bootstrap}
Jean-Bastien Grill, Florian Strub, Florent Altch{\'e}, Corentin Tallec, Pierre
  Richemond, Elena Buchatskaya, Carl Doersch, Bernardo Avila~Pires, Zhaohan
  Guo, Mohammad Gheshlaghi~Azar, et~al.
\newblock Bootstrap your own latent-a new approach to self-supervised learning.
\newblock {\em Advances in Neural Information Processing Systems},
  33:21271--21284, 2020.

\bibitem{xia2021self}
Xin Xia, Hongzhi Yin, Junliang Yu, Qinyong Wang, Lizhen Cui, and Xiangliang
  Zhang.
\newblock Self-supervised hypergraph convolutional networks for session-based
  recommendation.
\newblock In {\em Proceedings of the AAAI Conference on Artificial
  Intelligence}, volume~35, pages 4503--4511, 2021.

\bibitem{du2021hypergraph}
Boxin Du, Changhe Yuan, Robert Barton, Tal Neiman, and Hanghang Tong.
\newblock Hypergraph pre-training with graph neural networks.
\newblock {\em arXiv preprint arXiv:2105.10862}, 2021.

\bibitem{he2017neural}
Xiangnan He, Lizi Liao, Hanwang Zhang, Liqiang Nie, Xia Hu, and Tat-Seng Chua.
\newblock Neural collaborative filtering.
\newblock In {\em Proceedings of the 26th international conference on world
  wide web}, pages 173--182, 2017.

\bibitem{wei2021model}
Tianxin Wei, Fuli Feng, Jiawei Chen, Ziwei Wu, Jinfeng Yi, and Xiangnan He.
\newblock Model-agnostic counterfactual reasoning for eliminating popularity
  bias in recommender system.
\newblock In {\em Proceedings of the 27th ACM SIGKDD Conference on Knowledge
  Discovery \& Data Mining}, pages 1791--1800, 2021.

\bibitem{wei2022comprehensive}
Tianxin Wei and Jingrui He.
\newblock Comprehensive fair meta-learned recommender system.
\newblock In {\em Proceedings of the 28th ACM SIGKDD Conference on Knowledge
  Discovery and Data Mining}, pages 1989--1999, 2022.

\bibitem{zhang2021causal}
Yang Zhang, Fuli Feng, Xiangnan He, Tianxin Wei, Chonggang Song, Guohui Ling,
  and Yongdong Zhang.
\newblock Causal intervention for leveraging popularity bias in recommendation.
\newblock In {\em Proceedings of the 44th International ACM SIGIR Conference on
  Research and Development in Information Retrieval}, pages 11--20, 2021.

\bibitem{kingma2013auto}
Diederik~P Kingma and Max Welling.
\newblock Auto-encoding variational bayes.
\newblock {\em arXiv preprint arXiv:1312.6114}, 2013.

\bibitem{kipf2016variational}
Thomas~N Kipf and Max Welling.
\newblock Variational graph auto-encoders.
\newblock {\em arXiv preprint arXiv:1611.07308}, 2016.

\bibitem{kingma2019introduction}
Diederik~P Kingma, Max Welling, et~al.
\newblock An introduction to variational autoencoders.
\newblock {\em Foundations and Trends{\textregistered} in Machine Learning},
  12(4):307--392, 2019.

\bibitem{you2021bayesian}
Yuning You, Yue Cao, Tianlong Chen, Zhangyang Wang, and Yang Shen.
\newblock Bayesian modeling and uncertainty quantification for learning to
  optimize: What, why, and how.
\newblock In {\em International Conference on Learning Representations}, 2021.

\bibitem{zugner2019adversarial}
Daniel Z{\"u}gner and Stephan G{\"u}nnemann.
\newblock Adversarial attacks on graph neural networks via meta learning.
\newblock {\em arXiv preprint arXiv:1902.08412}, 2019.

\bibitem{zugner2018adversarial}
Daniel Z{\"u}gner, Amir Akbarnejad, and Stephan G{\"u}nnemann.
\newblock Adversarial attacks on neural networks for graph data.
\newblock In {\em Proceedings of the 24th ACM SIGKDD International Conference
  on Knowledge Discovery \& Data Mining}, pages 2847--2856, 2018.

\bibitem{asuncion2007uci}
Arthur Asuncion and David Newman.
\newblock Uci machine learning repository, 2007.

\bibitem{jordan2015effect}
Kareem~L Jordan and Tina~L Freiburger.
\newblock The effect of race/ethnicity on sentencing: Examining sentence type,
  jail length, and prison length.
\newblock {\em Journal of Ethnicity in Criminal Justice}, 13(3):179--196, 2015.

\bibitem{yeh2009comparisons}
I-Cheng Yeh and Che-hui Lien.
\newblock The comparisons of data mining techniques for the predictive accuracy
  of probability of default of credit card clients.
\newblock {\em Expert systems with applications}, 36(2):2473--2480, 2009.

\end{thebibliography}
% }
% \newpage

%%%%%%%%%%%%%%%%%%%%%%%%%%%%%%%%%%%%%%%%%%%%%%%%%%%%%%%%%%%%
\section*{Checklist}

%%% BEGIN INSTRUCTIONS %%%
The checklist follows the references.  Please
read the checklist guidelines carefully for information on how to answer these
questions.  For each question, change the default \answerTODO{} to \answerYes{},
\answerNo{}, or \answerNA{}.  You are strongly encouraged to include a {\bf
justification to your answer}, either by referencing the appropriate section of
your paper or providing a brief inline description.  For example:
\begin{itemize}
  \item Did you include the license to the code and datasets? \answerYes{See Section.}
  \item Did you include the license to the code and datasets? \answerNo{The code and the data are proprietary.}
  \item Did you include the license to the code and datasets? \answerNA{}
\end{itemize}
Please do not modify the questions and only use the provided macros for your
answers.  Note that the Checklist section does not count towards the page
limit.  In your paper, please delete this instructions block and only keep the
Checklist section heading above along with the questions/answers below.
%%% END INSTRUCTIONS %%%

\begin{enumerate}

\item For all authors...
\begin{enumerate}
  \item Do the main claims made in the abstract and introduction accurately reflect the paper's contributions and scope?
    \answerYes
  \item Did you describe the limitations of your work?
    \answerYes{Please check Apendix B.}
  \item Did you discuss any potential negative societal impacts of your work?
    \answerNA{We are not aware of any potential ethical issues regarding our work. More discussions can be found in Appendix B.}
  \item Have you read the ethics review guidelines and ensured that your paper conforms to them?
    \answerYes
\end{enumerate}

\item If you are including theoretical results...
\begin{enumerate}
  \item Did you state the full set of assumptions of all theoretical results?
    \answerNA
        \item Did you include complete proofs of all theoretical results?
    \answerNA
\end{enumerate}

\item If you ran experiments...
\begin{enumerate}
  \item Did you include the code, data, and instructions needed to reproduce the main experimental results (either in the supplemental material or as a URL)?
    \answerYes{We add them in the supplementary materials.}
  \item Did you specify all the training details (e.g., data splits, hyperparameters, how they were chosen)?
    \answerYes{We write the implementation details in Appendix C.}
        \item Did you report error bars (e.g., with respect to the random seed after running experiments multiple times)?
    \answerYes{We report the standard deviation of 20 runs with different splits.}
        \item Did you include the total amount of compute and the type of resources used (e.g., type of GPUs, internal cluster, or cloud provider)?
    \answerYes{We specify the computing resouce in Appendix C.}
\end{enumerate}

\item If you are using existing assets (e.g., code, data, models) or curating/releasing new assets...
\begin{enumerate}
  \item If your work uses existing assets, did you cite the creators?
    \answerYes
  \item Did you mention the license of the assets?
    \answerYes{We discuss it in Appendix C.}
  \item Did you include any new assets either in the supplemental material or as a URL?
    \answerYes{We submit the code in the supplementary materials.}
    % {All our used methods and datasets are introduced in the main text. Our code will be directly added in the Appendix.}
  \item Did you discuss whether and how consent was obtained from people whose data you're using/curating?
    \answerYes{We discuss it in Appendix C.}
  \item Did you discuss whether the data you are using/curating contains personally identifiable information or offensive content?
    \answerYes{The data sets we use don't contain any personally identifiable information.}
\end{enumerate}

\item If you used crowdsourcing or conducted research with human subjects...
\begin{enumerate}
  \item Did you include the full text of instructions given to participants and screenshots, if applicable?
    \answerNA{We don't conduct research with human subjects.}
  \item Did you describe any potential participant risks, with links to Institutional Review Board (IRB) approvals, if applicable?
    \answerNA
  \item Did you include the estimated hourly wage paid to participants and the total amount spent on participant compensation?
    \answerNA
\end{enumerate}

\end{enumerate}

%%%%%%%%%%%%%%%%%%%%%%%%%%%%%%%%%%%%%%%%%%%%%%%%%%%%%%%%%%%%

\end{document}